\documentclass[english,smallextended]{article}
\usepackage[latin9]{inputenc}
\usepackage{geometry}
\usepackage{array}
\usepackage{multirow}
\usepackage{amsmath}
\usepackage{amssymb}
\usepackage{graphicx}
\usepackage{microtype}

\makeatletter

\providecommand{\tabularnewline}{\\}

\date{}

\makeatother

\usepackage{babel}
\begin{document}

\title{Relational dynamic memory networks}

\author{Trang Pham, Truyen Tran and Svetha Venkatesh\\
Applied AI Institute, Deakin University, Australia\\
\emph{\{phtra,truyen.tran,svetha.venkatesh\}@deakin.edu.au} }

\maketitle
\global\long\def\Model{\text{RDMN}}
\global\long\def\OurModel{\Model}

\begin{abstract}
Neural networks excel in detecting regular patterns but are less successful
in representing and manipulating complex data structures, possibly
due to the lack of an external memory. This has led to the recent
development of a new line of architectures known as Memory-Augmented
Neural Networks (MANNs), each of which consists of a neural network
that interacts with an external memory matrix. However, this RAM-like
memory matrix is unstructured and thus does not naturally encode structured
objects. Here we design a new MANN dubbed Relational Dynamic Memory
Network ($\Model$) to bridge the gap. Like existing MANNs, $\Model$
has a neural controller but its memory is structured as multi-relational
graphs. $\Model$ uses the memory to represent and manipulate graph-structured
data in response to query; and as a neural network, $\Model$ is trainable
from labeled data. Thus $\Model$ learns to answer queries about a
set of graph-structured objects without explicit programming. We evaluate
the capability of $\Model$ on several important prediction problems,
including software vulnerability, molecular bioactivity and chemical-chemical
interaction. Results demonstrate the efficacy of the proposed model. 

\end{abstract}

\global\long\def\xb{\boldsymbol{x}}
\global\long\def\yb{\boldsymbol{y}}
\global\long\def\eb{\boldsymbol{e}}
\global\long\def\zb{\boldsymbol{z}}
\global\long\def\hb{\boldsymbol{h}}
\global\long\def\ab{\boldsymbol{a}}
\global\long\def\bb{\boldsymbol{b}}
\global\long\def\cb{\boldsymbol{c}}
\global\long\def\sigmab{\boldsymbol{\sigma}}
\global\long\def\gammab{\boldsymbol{\gamma}}
\global\long\def\alphab{\boldsymbol{\alpha}}
\global\long\def\betab{\boldsymbol{\beta}}
\global\long\def\rb{\boldsymbol{r}}
\global\long\def\gb{\boldsymbol{g}}
\global\long\def\Deltab{\boldsymbol{\Delta}}
\global\long\def\wb{\boldsymbol{w}}
\global\long\def\vb{\boldsymbol{v}}
\global\long\def\eb{\boldsymbol{e}}
\global\long\def\sb{\boldsymbol{s}}
\global\long\def\ub{\boldsymbol{u}}
\global\long\def\fb{\boldsymbol{f}}
\global\long\def\mb{\boldsymbol{m}}
\global\long\def\qb{\boldsymbol{q}}
\global\long\def\thetab{\boldsymbol{\theta}}
\global\long\def\Mb{\boldsymbol{M}}
\global\long\def\Ab{\boldsymbol{A}}
\global\long\def\kb{\boldsymbol{k}}

\section{Introduction}

To support reasoning -- the process of forming answer to a new question
by deliberately manipulating previously acquired knowledge \cite{bottou2014machine}
-- intelligent systems need a working memory to load, hold, integrate
and alter information needed for processing \cite{diamond2013executive}.
This has inspired a recent line of research collectively known as
memory-augmented neural networks (MANNs), in which a neural network
reads from and writes to an external memory matrix \cite{graves2016hybrid,kumar2016ask,sukhbaatar2015end}.
This results in a powerful differentiable machinery that can learn
programs from data and answer complex queries. Technically, memory
provides a short-cut for passing signal and gradient between query,
input and output \cite{le2018dual}, making credit-assignment easier
in a long chain of computation. However, these memory modules have
been developed to be generic without considering the structural information
that may be available in the objects being queried about. For structured
objects such as graphs, encoding structural information into a flat
memory matrix is not straightforward \cite{graves2016hybrid}. We
conjecture that a memory architecture that is reflective of the structure
of the data might be easier to train and generate a more focused answer.

Here we introduce a new MANN called Relational Dynamic Memory Networks
($\Model$) capable of answering queries about structured data\footnote{A preliminary version of this work was published in a conference \cite{pham2018graph}.}.
We consider the case where data are attributed graphs, e.g., molecular
graphs or function call graphs in software. $\OurModel$ is composed
of a controller and a memory dynamically organized as a set of networks
of memory cells. The memory structure is not fixed but shaped by the
structure within the input data conditioned on the query. This memory
design is partly inspired by the current understanding of working
memory as dynamic networks emerged from functional coordination between
active regions in brain \cite{braun2015dynamic,eriksson2015neurocognitive,stokes2015activity}.
The query triggers the controller in $\Model$ to initiate a reasoning
episode, during which the controller first prepares the memory structure
and content based on the input data, then iteratively manipulates
the memory states until a probable answer is reached. Neighboring
memory cells interact directly during the reasoning process, i.e.,
cell states are updated by aggregating the write content from the
controller as well as the messages sent by neighboring cells. Distant
cells within a network or between networks interact indirectly through
the exchanging messages with the controller.

With this architecture, the $\OurModel$ supports learning to answer
queries about not just a single graph, but also several interacting
graphs. In other words, $\Model$ learns o solve problems of the form
\texttt{(query, \{set of graphs\}, ?)}. This is a generalized form
of \emph{graph-graph interaction}, an under-explored machine learning
area in its own right. An example of query over single graph is predicting
whether a drug molecule (a graph) has any positive effect on a type
of disease (a query). An example of querying over multiple graphs
is chemical-chemical interaction prediction, where the query can include
environmental factors, the graphs are basic molecular structures,
and the answer is the interaction strength between molecules \cite{fooshee2018deep,kwon2018end}. 

In summary, we claim the following contributions:
\begin{itemize}
\item A novel differentiable architecture named Relational Dynamic Memory
Networks ($\Model$), which consists of a structured working memory
module augmented to a neural network. 
\item A general solution for learning to answer queries about multiple graphs.
In particular, it solves a relatively novel problem of predicting
graph-graph interaction.
\item Validation of $\OurModel$ on three distinct tasks: software source
code vulnerability assessment, molecular bioactivity prediction, and
chemical-chemical interaction prediction.
\end{itemize}
The rest of the paper is organized as follows. Section~\ref{sec:Prelim}
briefly introduces preliminaries on memory-augmented neural networks
and graph neural networks. The main contribution of the paper, the
$\Model$, is described in Section~\ref{sec:Graph-memory-networks}
with implementation detailed in Section~\ref{sec:Implementation}.
Experiments and results are reported in Section~\ref{sec:Experiments-and-results}.
Section~\ref{sec:Related-work} reviews related work, followed by
conclusion.

\section{Preliminaries \label{sec:Prelim}}

\subsection{Memory-augmented neural nets \label{subsec:Memory-augmented-neural-nets}}

A MANN consists of a neural controller augmented with an external
memory. The controller can be a feedforward net (memoryless) or a
recurrent net (which has its own short-term memory). The external
memory is often modeled as a matrix. At each time step, the controller
reads an input, updates the memory, and optionally emits an output.
With stationary update rules, the memory can be rolled out over time
into a recurrent matrix net \cite{do2018learning}. Let us denote
the memory matrix as $\Mb\in\mathbb{R}^{d\times m}$ for $d$ dimensions
and $m$ slots. Upon seeing a new evidence $\xb$ (which could be
empty), the controller generates a write vector $\wb\in\mathbb{R}^{p}$
which causes a candidate memory update. As an example, the update
may take the following form:
\begin{equation}
\tilde{\Mb}\leftarrow\phi\left(W\wb\mathbf{1}^{\top}+U\Mb R+B\right)\label{eq:matrix-rnn}
\end{equation}
where $W\in\mathbb{R}^{d\times p}$ is data encoding matrix, $U\in\mathbb{R}^{d\times d}$
is transition matrix , $R\in\mathbb{R}^{m\times m}$ is the graph
of relation between memory slots, and $B\in\mathbb{R}^{d\times m}$
is bias. The memory is then updated as: $\Mb\leftarrow f\left(\Mb,\tilde{\Mb}\right)$
for some function $f$. For example, a simple but effective update
is linear forgetting: $f\left(\Mb,\tilde{\Mb}\right)=\alphab*\Mb+(1-\alphab)*\tilde{\Mb}$,
where $*$ is element-wise multiplication, and $\alphab\in(0,\boldsymbol{1})$
is learnable forget-gate.

The controller then reads the memory to determine what to do next.
For example, the End-to-End Memory Network \cite{sukhbaatar2015end}
maintains a dynamic state of an answer $\ub$ by reading the memory
as follows:

\begin{align*}
\ab & \leftarrow\text{softmax}\left(\Mb^{\top}A\ub\right)\\
\ub & \leftarrow U\ub+BM\ab
\end{align*}
where $A,U,B$ are trainable parameters. Here $\kb=A\ub$ plays the
role of a key in the content-addressing scheme, and $\ab$ assigns
an attention weight to each memory slot.

\subsection{Graph neural networks \label{subsec:Graph-neural-networks}}

A graph is a tuple $\text{\textbf{G}=\{\textbf{A}, \textbf{R}, \textbf{X}\}}$,
where $\text{\textbf{A}}=\left\{ a^{1},...,a^{M}\right\} $ are $M$
nodes; $\text{\textbf{X}}=\left\{ \xb^{1},...,\xb^{M}\right\} $ is
the set of node features; and $\text{\textbf{R}}$ is the set of relations
in the graph. Each tuple $\left\{ a^{i},a^{j},r,\bb^{ij}\right\} \in\text{\textbf{R}}$
describes a relation of type $r$ ($r=1...R$) between two nodes $a^{i}$
and $a^{j}$. The relations can be unidirectional or bi-directional.
The vector $\bb^{ij}$ represents the link features. Node $a^{j}$
is a neighbor of $a^{i}$ if there is a connection between the two
nodes. Let $\mathcal{N}(i)$ be the set of all neighbors of $a^{i}$
and $\mathcal{N}_{r}(i)$ be the set of neighbors connected to $a^{i}$
through type $r$. This implies $\mathcal{N}(i)=\cup_{r}\mathcal{N}_{r}(i)$.

Graph neural networks are a class of neural nets that model the graph
structure directly. The most common type is message passing graph
neural networks \cite{gilmer2017neural,pham2017column,scarselli2009graph},
which update note states using messages sent from the neighborhood.
In the \textit{message aggregation} step, we combine multiple messages
sent to node $i$ into a single message vector $\mb_{i}$:

\begin{equation}
\mb_{i}^{t}=g^{\text{a}}\left(\xb_{i}^{t-1},\left\{ \left(\xb_{j}^{t-1},\eb_{ij}\right)\right\} _{j\in\mathcal{N}(i)}\right)\label{eq:message_aggregation}
\end{equation}
where $g^{\text{a}}(\cdot)$ can be an attention \cite{bahdanau2014neural}
or a pooling architecture. During the \textit{state update} step,
the node state is updated as follows:

\begin{equation}
\xb_{i}^{t}\leftarrow g^{\text{u}}\left(\xb_{i}^{t-1},\mb_{i}^{t}\right)\label{eq:node_update}
\end{equation}
where $g^{\text{u}}(\cdot)$ can be any type of deep neural networks
such as MLP \cite{kipf2016semi}, RNN \cite{scarselli2009graph},
GRU \cite{li2016gated} or Highway Net \cite{pham2017column}.

\section{Relational dynamic memory \label{sec:Graph-memory-networks}}

\begin{figure}[h]
\begin{centering}
\includegraphics[width=0.9\textwidth]{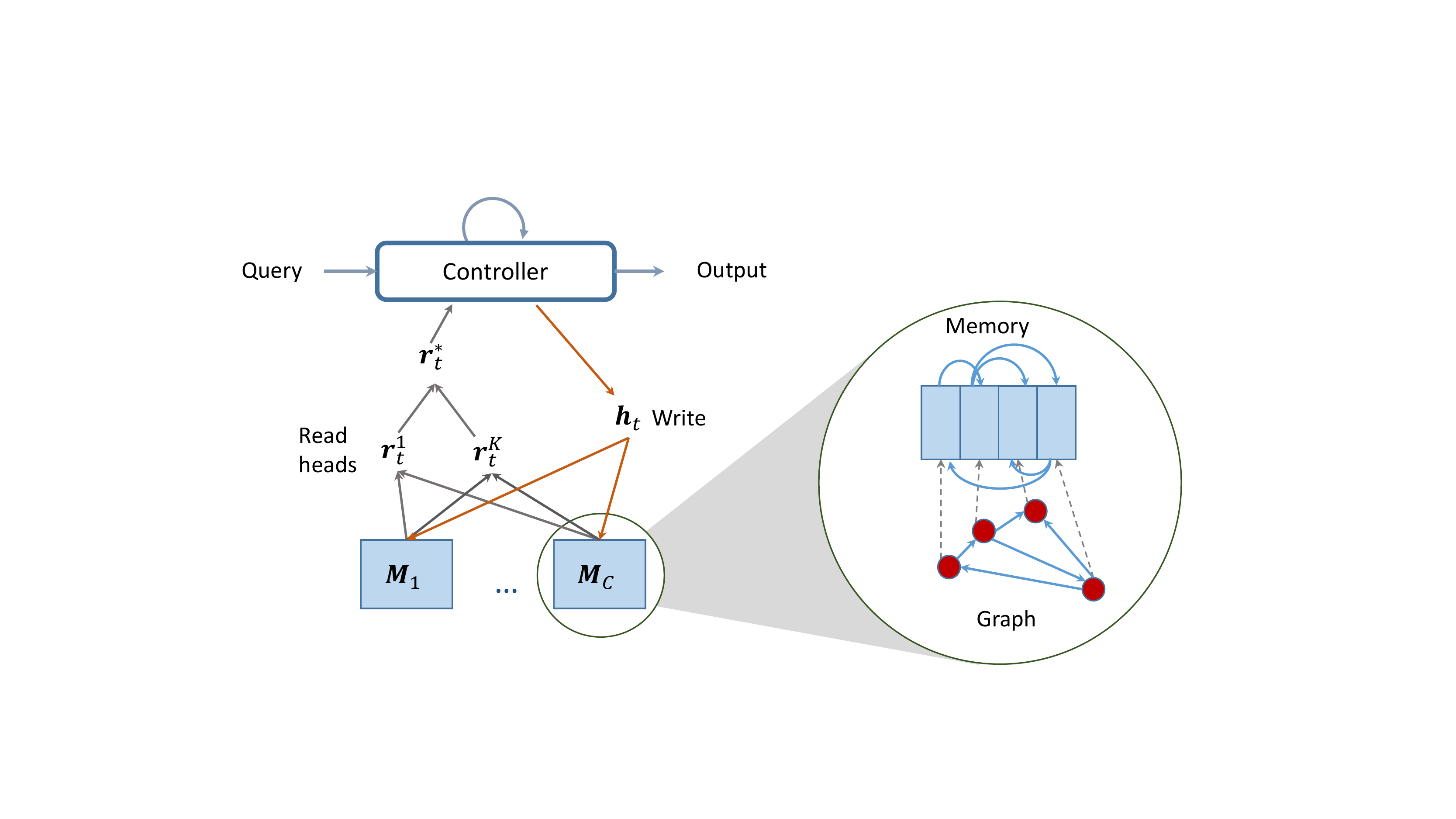}
\par\end{centering}
\centering{}\caption{Relational Dynamic Memory Networks ($\protect\Model$). There are
multiple structured memory components $\left\{ \protect\Mb_{c}\right\} _{c=1}^{C}$,
each of which is associated with an input graph. When a query is presented,
the controller reads the query and intializes the memory. For each
graph, each node is embedded into a memory cell. Then during the reasoning
process, the controller iteratively reads from and writes to the memory.
Finally, the controller emits the output. \label{fig:Model}}
\end{figure}

Here we present our main contribution, the Relational Dynamic Memory
Network ($\Model$), which is built upon two independent concepts,
the Memory-Augmented Neural Net (see Section~\ref{subsec:Memory-augmented-neural-nets})
and Graph Neural Net (see Section~\ref{subsec:Graph-neural-networks}).
$\Model$ is a neural computer, similar to Neural Turing Machines
(NTM) \cite{graves2014neural,graves2016hybrid}, thus consisting of
a CPU-like controller and a RAM-like memory. The controller is a recurrent
neural network responsible for memory manipulation. Like NTMs, $\Model$
is fully differentiable allowing it to be trained end-to-end using
gradient-based techniques without explicit programming. However, unlike
NTMs which have flat memory and thus make no assumption about the
data structure other than the sequential arrival, $\Model$ has explicitly
structured memory suitable for graph data. In $\Model$, the memory
module can have one or multiple components (Fig.~\ref{fig:Model}),
each of which is a graph of cells. Links between cells are dynamically
defined by the structure of the input data and the query.

Presented with input graphs and a query, $\Model$ initiates a reasoning
episode, during which the machine dynamically allocates memory to
represent the input graphs, with one memory component per graph. The
memory structure of a component reflects the corresponding input graph
with respect to the query. Then the controller takes the query as
the input and repeatedly reads from the memory, processes and writes
back to the memory cells. At each reasoning step, the cell content
is updated by the signals from the controller and its neighbor memory
cells. Throughout the reasoning episode, the memory cells evolve from
the original input to a refined stage, preparing the controller for
generating the output. The interaction between memory components is
mediated through message exchanging with the controller. For $C$
memory components, the system, when rolled out during reasoning, consists
of $C+1$ recurrent \emph{matrix} neural networks interacting with
each other \cite{do2018learning}.

\subsection{Encoder}

The encoder prepares data for the reasoning process, i.e., encoding
queries into the controller and data into the memory. The query $q$
is first encoded into a vector as:
\begin{equation}
\qb=\text{q\_encode}\left(q\right)\label{eq:q_encode}
\end{equation}

Let $X_{c}$ be a graph-structured representation of object $c$,
for $c=1,2,...,C$. The representation $X_{c}$ is loaded into a memory
matrix $\Mb_{c}$:
\begin{equation}
\left(\Mb_{c},\Ab_{c}\right)=\text{m\_load}\left(X_{c},q\right)\label{eq:m_load}
\end{equation}
where $\Ab_{c}$ is relational structure between memory slots. The
dependency on the query $q$ offers flexibility to define query-specific
memory, which could be more economical in cases where the query provides
constraints on the relevant parts of the object.

Assume that the query and object $c$ together impose a set $\mathcal{R}_{c}$
of pairwise relations between memory cells. More precisely, any pair
of cells will have zero or more relations draw from the relation set.
For example, for $c$ is a molecule, then $\Mb_{c}[i]$ can be the
(learnable) embedding of the atom at node $i$ of the molecular graph,
and the relations can be bonding types (e.g., ion or valence bonds).
The relations can be modeled as a collection of adjacency matrices,
that is, $\Ab_{c}=\left\{ A_{c,r}\right\} _{r\in\mathcal{R}_{c}}$.

Given $\qb$ and $\Mb=\left\{ \Mb_{1},\Mb_{2},...,\Mb_{C}\right\} $,
the controller initiates a reasoning episode to compute the output
probability $P_{\theta}\left(\yb\mid\Mb,\qb\right)$. In what follows,
we briefly present how the computation occurs, leaving the detailed
implementation in Section~\ref{sec:Implementation}.

\subsection{Reasoning processes}

The controller manages the reasoning process by updating its own state
and the working memory. Let $\hb_{t}$ be the state of the controller
at time $t$ $\left(t=0,...,T\right)$. At time $t=0$, the state
is initialized as $\hb_{0}=\qb$. At $t=1,2,...,T$, the controller
retrieves a content $\rb_{t}^{*}$ from the memory at time $t$ as
follows:

\begin{align}
\rb_{c,t}^{k} & =\text{m\_read}_{k}\left(\Mb_{c,t-1},\hb_{t-1}\right)\label{eq:m_read}\\
\bar{\rb}_{t}^{k} & =\text{r\_aggregate}\left\{ \rb_{c,t}^{k}\right\} _{c=1}^{C}\label{eq:r_aggregate}\\
\rb_{t}^{*} & =\text{r\_combine}\left\{ \bar{\rb}_{t}^{k}\right\} _{k=1}^{K}\label{eq:r_combine}
\end{align}
where $k=1,2,...,K$ denotes the index of read head. The use of multiple
read heads accounts for multiple pieces of information that may be
relevant at any step $t$. The $\text{r\_aggregate\ensuremath{\left\{  \cdot\right\} } }$
function operates on the variable-size set of retrieved memory vectors
and combines them into one vector. The $\text{r\_combine\ensuremath{\left\{  \cdot\right\} } }$function
operates on the fixed-size set of vectors returned by the read-heads.

Once the content $\rb_{t}^{*}$ has been read, the controller state
is updated as:

\begin{align}
\hb_{t} & =\text{s\_update}\left(\hb_{t-1},\rb_{t}^{*}\right)\label{eq:s_update}
\end{align}
This initiates an update of the memory:

\begin{align}
\Mb_{c,t} & =\text{m\_update}\left(\Mb_{c,t-1},\hb_{t},\Ab_{c}\right)\label{eq:m_update}
\end{align}
for $c=1,2,...,C$.

\subsection{Decoder}

At the end of the reasoning process, the controller predicts an output,
that is
\begin{equation}
\yb=\text{decode}\left(\hb_{T},\qb\right)\label{eq:decode}
\end{equation}
The decoder is a task-specific model, which could be a deep feedforward
net for vector output, or a RNN for sequence output.

\section{An implementation \label{sec:Implementation}}

We now present a specific realization of the generic framework proposed
in Section~\ref{sec:Graph-memory-networks} and validated in Section~\ref{sec:Experiments-and-results}.
All operators are parameterized and differentiable. We use ReLU units
for all steps and Dropout is applied at the first and the last steps
of the controller and the memory cells.

\subsection{Operators}
\begin{itemize}
\item The \textbf{q\_encode()} operator in Eq.~(\ref{eq:q_encode}) translates
the query $q$ into a vector $\qb$. This could be simply a look-up
table when $q$ is discrete (e.g., task or word). For textual query,
we can use an RNN to represent $q$, and $\qb$ is the last (or averaged)
state of the RNN.
\item The \textbf{m\_load()} operator in Eq.~(\ref{eq:m_load}), $\left(\Mb_{c},\Ab_{c}\right)=\text{m\_load}\left(X_{c},q\right)$,
returns both the memory matrix $\Mb_{c}$ and the relational structure
$\Ab_{c}$ given the query $q$. This is problem dependent. One example
is when $X_{c}$ represents a molecule, and $q$ specifies a sub-structure
of the molecule. Each atom $i$ in the substructure is embedded into
a vector $\mb_{i}$, which is first assigned to the memory cell $\Mb_{c}[i]$.
The relational structure $\Ab_{c}$ reflects the bonding between the
atoms in the substructure.
\item For the \textbf{m\_read()} operator in Eq.~(\ref{eq:m_read}), a
content-based addressing scheme, also known as soft attention, is
employed. At each time step $t$ $\left(t=1,2,...,T\right)$, the
read vector $\rb_{c,t}^{k}$ returned by the reading head $k$ ($k=1,2,...,K)$
over memory component $c$ $\left(c=1,2,...,C\right)$ is a sum of
all memory cells, weighted by the attention probability vector $\ab$:
\begin{eqnarray*}
\ab & = & \text{attention}_{k}\left(\Mb_{c,t-1},\hb_{t-1}\right)\\
\rb_{t} & = & \Mb_{c,t-1}\ab
\end{eqnarray*}
where $\text{attention}_{k}()$ is implemented as follows:
\begin{eqnarray*}
\ab[i] & = & \text{softmax}\left(\vb_{k}^{\top}\text{tanh}\left(W_{a,k}\Mb_{c,t-1}[i]+U_{a,k}\hb_{t-1}\right)\right).
\end{eqnarray*}
\item The controller is implemented as a recurrent net, where \textbf{s\_update()}
operator in Eq.~(\ref{eq:s_update}) reads:
\begin{align}
\tilde{\hb}_{t} & =g\left(W_{h}\hb_{t-1}+U_{h}\rb_{t}\right)\nonumber \\
\hb_{t} & =\alphab_{h}*\hb_{t-1}+\left(1-\alphab_{h}\right)*\tilde{\hb}_{t}
\end{align}
where $(*)$ is element-wise multiplication; and $\alphab_{h}\in(\boldsymbol{0},\boldsymbol{1})$
is trainable forgetting gate that moderates the amount of information
flowing from the previous step, e.g., $\alphab_{h}=\sigma\left(\text{nnet}\left[\hb_{t-1},\rb_{t}\right]\right)$
for sigmoid function $\sigma$. 
\item For the \textbf{m\_update()} operator in Eq.~(\ref{eq:m_update}),
the memory is updated as follows:
\begin{eqnarray}
\tilde{\Mb}_{c,t} & = & g\left(U_{m}\hb_{t}\boldsymbol{1}^{\top}+W_{m}\Mb_{c,t-1}+\sum_{r}V_{c,r}\Mb_{c,t-1}\hat{\Ab}_{c,r}\right)\nonumber \\
\Mb_{c,t}\left[i\right] & = & \alphab_{c,i}*\Mb_{c,t-1}\left[i\right]+\left(1-\alphab_{c,i}\right)*\tilde{\Mb}_{c,t}\left[i\right]
\end{eqnarray}
for $c=1,2,...,C$; where $\hat{\Ab}_{c,r}$ is the normalized adjacency
matrix, i.e., $\hat{\Ab}_{c,r}\left[i,j\right]=\frac{\Ab_{c,r}\left[i,j\right]}{\sum_{i}\Ab_{c,r}\left[i,j\right]}$;
$(*)$ is element-wise multiplication; and $\alphab_{c,i}\in(\boldsymbol{0},\boldsymbol{1})$
is a trainable forgetting gate. 
\item The \textbf{r\_aggregate()} and the \textbf{r\_combine()} operators
in Eqs.~(\ref{eq:r_aggregate},\ref{eq:r_combine}) are simply averaging,
i.e.,
\begin{align*}
\bar{\rb}_{t}^{k} & =\frac{1}{C}\sum_{c=1}^{C}\rb_{c,t}^{k};\quad\quad\rb_{t}^{*}=\frac{1}{K}\sum_{k=1}^{K}\bar{\rb}_{t}^{k}.
\end{align*}
\end{itemize}

\subsection{$\protect\Model$ for multi-task learning}

$\Model$ can be easily applied for multi-task learning. Suppose that
the dataset contains $n$ tasks. We can use the query to indicate
the task. If a graph is from task $k$, the query for the graph is
an one-hot vector of size $n$: $\qb=[0,0,...,1,0,...]$, where $\qb^{k}=1$
and $\qb^{j}=0$ for $j=1,...,n,\text{ }j\neq k$. The task index
now becomes the input signal for $\Model$. With the signal from the
task-specific query, the attention can identify which substructure
is important for a specific task to attend on.

\subsection{Training}

Given a training set, in which each data instance has the form \texttt{(query,
\{graph$_{1}$, graph$_{2}$, ..., graph$_{C}$\}, answer)}, $\Model$
is trained in a supervised fashion from end-to-end, i.e., by minimizing
a loss function. The typical loss is $-\log P\left(\yb\mid q,\left\{ X_{c}\right\} _{c=1}^{C}\right)$.
As the entire system is differentiable, gradient-based techniques
can be applied for parameter optimization\footnote{Discrete operators can also be implemented with help of policy gradient
algorithms.}.

\section{Experiments and results \label{sec:Experiments-and-results}}

In this section, we demonstrate $\OurModel$ on three applications:
software vulnerability detection (single query, Section~\ref{subsec:Query:-Software-vulnerability}),
molecular activity prediction (multiple queries, Section~\ref{subsec:Query:-Molecular-activities}),
and chemical-chemical interaction (query about graph-graph interaction,
Section~\ref{subsec:Query:-Chemical-reaction}).

\subsection{Software vulnerability \label{subsec:Query:-Software-vulnerability}}

This application asks if a piece of source code is potentially vulnerable
to security risks. In particular, we consider each Java class as a
piece, which consists of attribute declarations and methods. A class
is then represented as a graph, where nodes are methods and edges
are function calls between methods (thus, there is only one relation
type). Class-level declaration is used as query, and the answer is
binary indicator of vulnerability. The dataset was collected from
\cite{dam2017automatic}, which consists of 18 Java projects. The
dataset is pre-processed by removing all replicated files of different
versions in the same projects. This results in 2,836 classes of which
1,020 are potentially vulnerable.

Embedding of graph nodes and query is pretrained as follows. Methods
and attribute declarations of Java classes are treated as sequences
of code tokens and their representation is learned through language
modeling using LSTM, that is to predict the next token $w_{t}$ given
previous tokens $\wb_{1:t-1}$ via $P\left(w_{t}\mid\wb_{1:t-1}\right)$.
Rare tokens are collectively designated as \texttt{\textless UNK\textgreater}.
The feature vector of each sequence is the mean of all hidden states
outputted by the LSTM. After this step, each sequence is represented
as a feature vector of 128 units. 

\begin{figure}[h]
\centering{}\includegraphics[viewport=100bp 0bp 550bp 400bp,clip,width=0.7\textwidth]{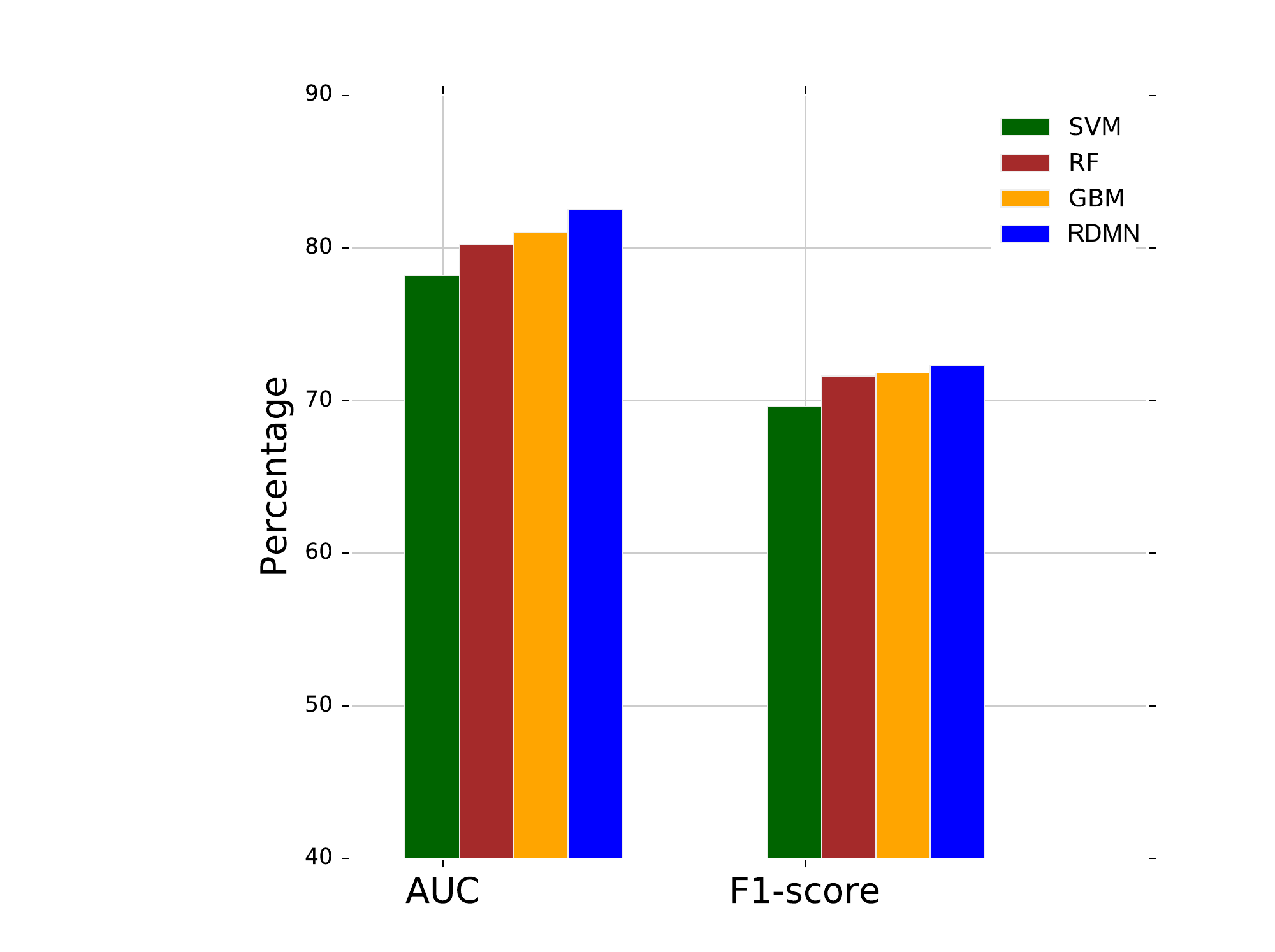}\caption{Performance on code vulnerability detection, measured in AUC and F1-score
(\%). Best viewed in color.\label{fig:Performance-on-Code}}
\end{figure}

For comparison, we implemented three strong non-neural classifiers:
Support Vector Machine (SVM), Random Forest (RF) and Gradient Boosting
Machine (GBM) running on the averaged feature vector of all methods
and attribute declarations. Fig.~\ref{fig:Performance-on-Code} reports
the performance measured by AUC (Area under the ROC curve) and F1-score
for $\Model$ and these three classifiers. $\Model$ is competitive,
although the improvement over the best performing method (GBM) is
not large (2 points on AUC and 1 point on F1-score).

\subsection{Molecular bioactivities \label{subsec:Query:-Molecular-activities}}

This application asks if a chemical compound is biologically active
on a given disease. Here each compound molecule is represented as
a graph, where nodes are atoms and edges are bond types between them.
Activity and disease form a query (e.g., coded as an one-hot vector).
The answer will be binary indicator of the activity with respect to
the given disease.

\paragraph{Datasets}

We conducted experiments on nine NCI BioAssay activity tests collected
from the PubChem website\footnote{https://pubchem.ncbi.nlm.nih.gov/}.
Seven of them are activity tests of chemical compounds against different
types of cancer: breast, colon, leukemia, lung, melanoma, central
nerve system and renal. The others are AIDS antiviral assay and Yeast
anticancer drug screen. Each BioAssay test contains records of activities
for chemical compounds. We chose the two most common activities for
classification: ``active'' and ``inactive''. The statistics of
data is reported in Table~\ref{tab:data-statistics}. The datasets
are listed by the ascending order of number of active compounds. ``\#
Graph'' is the number of graphs and ``\# Active'' is the number
of active graph against a BioAssay test These datasets are unbalanced,
therefore ``inactive'' compounds are randomly removed so that the
Yeast Anticancer dataset has 25,000 graphs and each of the other datasets
has 10,000 graphs.

\begin{table}[h]
\centering{}\caption{Summary of 9 NCI BioAssay datasets.\label{tab:data-statistics}}
\begin{tabular}{cccc}
\hline 
No. & Dataset & \# Active & \# Graph\tabularnewline
\hline 
1 & AIDS Antiviral & 1513 & 41,595\tabularnewline
2 & Renal Cancer & 2,325 & 41,560\tabularnewline
3 & Central Nervous System  & 2,430 & 42,473\tabularnewline
4 & Breast Cancer & 2,490 & 29,117\tabularnewline
5 & Melanoma & 2,767 & 39,737\tabularnewline
6 & Colon Cancer & 2,766 & 42,130\tabularnewline
7 & Lung Cancer & 3,026 & 38,588\tabularnewline
8 & Leukemia & 3,681 & 38,933\tabularnewline
9 & Yeast Anticancer & 10,090 & 86,130\tabularnewline
\hline 
\end{tabular}
\end{table}

We used RDKit\footnote{http://www.rdkit.org/} to extract the structure
of molecules, the atom and the bond features. An atom feature vector
is the concatenation of the one-hot vector of the atom and other features
such as atom degree and number of H atoms attached. We also make use
of bond features such as bond type and a binary value indicating if
a bond is in a ring.

\subsubsection{Experiment settings}

\paragraph{Baselines}

For comparison, we use several baselines:
\begin{itemize}
\item \emph{Classic classifiers running on feature vectors extracted from
molecular graphs}. Classifiers are Support Vector Machine (SVM), Random
Forest (RF), Gradient Boosting Machine (GBM), and multi-task neural
network (MT-NN) \cite{ramsundar2015massively}. For feature extraction,
following the standard practice in the computational chemistry literature,
we use the RDKit to extract molecular fingerprints -- the encoding
of the graph structure of the molecules by a vector of binary digits,
each presents the presence or absence of particular substructures
in the molecules. There are different algorithms to achieve molecular
fingerprints and the state of the art is the extended-connectivity
circular fingerprint (ECFP) \cite{rogers2010extended}. The dimension
of the fingerprint features is set by 1,024. SVM, RF and GBM predict
one activity at a time, but MT-NN predict all activities at the same
time.
\item \emph{Neural graphs which learn to extract features}. In particular,
we use Neural Fingerprint (NeuralFP) \cite{duvenaud2015convolutional}.
This predicts one activity at a time.
\end{itemize}

\paragraph{Model setting}

We set the number of hops to $T=10$ following recommendation in \cite{pham2016faster}.
Other hyper-parameters are tuned on the validation dataset.

\subsubsection{Results}

\paragraph{The impact of joint training}

\begin{figure}[h]
\centering{}\includegraphics[viewport=40bp 0bp 655bp 470bp,clip,width=0.8\textwidth]{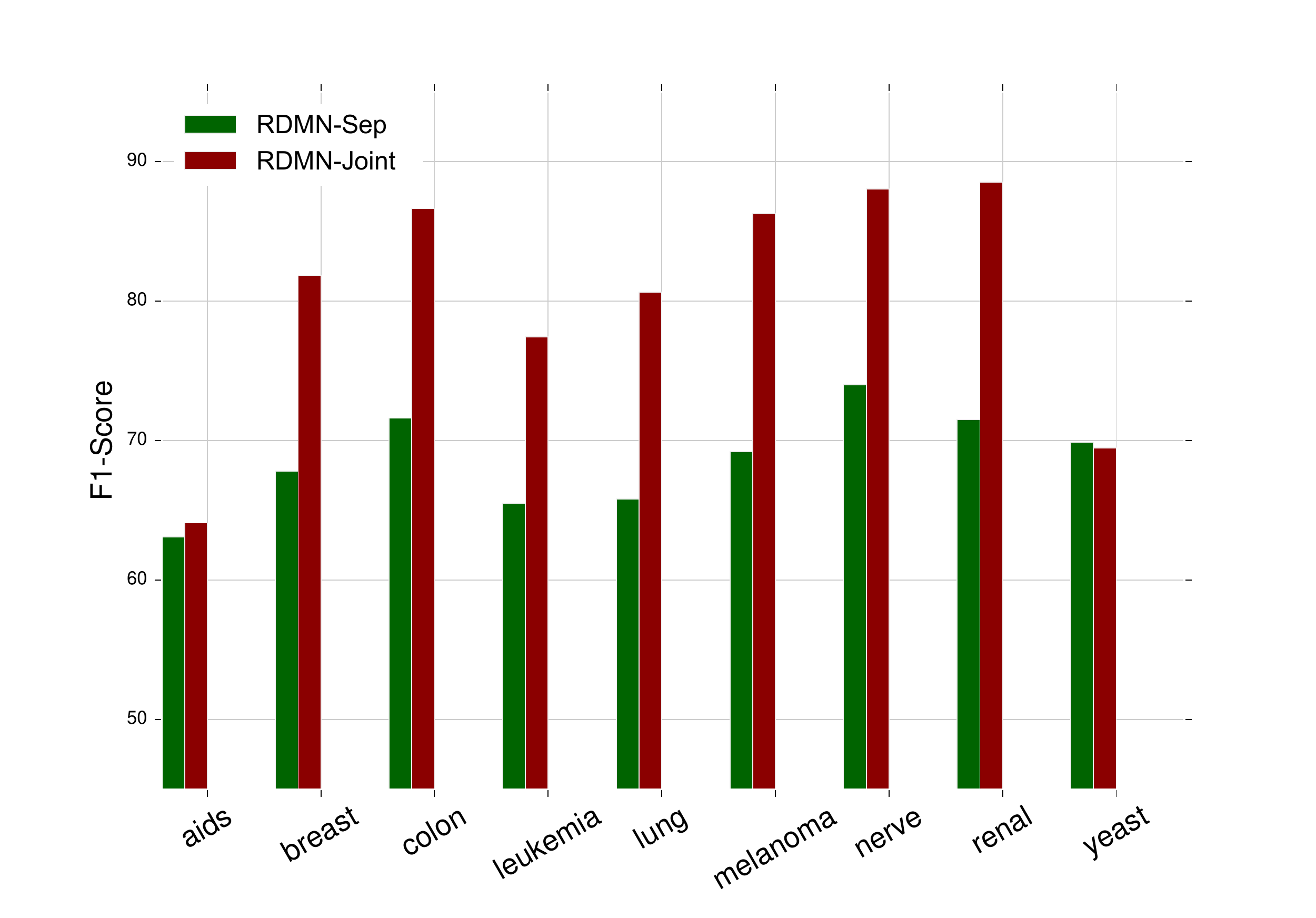}\caption{The comparison in performance of $\protect\Model$ when training separately
($\protect\Model$-Sep) and jointly ($\protect\Model$-Joint) for
all datasets. Best viewed in color.\label{fig:F1-score-each-dataset}}
\end{figure}

We have options to train each task separately (query set to unity,
one memory per task) or to train jointly (query is an one-hot vector,
shared memory for all tasks). To investigate more on how multi-task
learning impacts the performance of each task, we reports the F1-score
of $\Model$ in both separate and joint training settings on each
of the nine datasets (Fig.~\ref{fig:F1-score-each-dataset}). Joint
training with $\Model$ model improves the performance of seven datasets
on different types of cancers by 10\%-20\% on each task. However,
it has little effect on AIDS antiviral and Yeast anticancer datasets.
These could be explained by the fact that cancers share similar genomic
footprints and thus each cancer can borrow the strength of statistics
of the cancer family. This does not hold for the other conditions,
which are very different from the rest. However, joint training is
still desirable because we need to maintain just one model for all
queries.

\paragraph{The impact of more tasks}

We evaluate how the performance of $\Model$ on a particular dataset
is affected by the number of tasks. We chose AIDS antiviral, Breast
Cancer and Colon Cancer as the experimental datasets. For each experimental
dataset, we start to train it and then repeatedly add a new task and
retrain the model. The orders of the first three new tasks are: (AIDS,
Breast, Colon) for AIDS antiviral dataset, (Breast, AIDS, Colon) for
Breast Cancer dataset and (Colon, AIDS, Breast) for Colon Cancer dataset.
The orders of the remaining tasks are the same for three datasets:
(Leukemia, Lung, Melanoma, Nerve, Renal and Yeast).

\begin{figure}[h]
\centering{}%
\begin{tabular}{cc}
\includegraphics[viewport=10bp 0bp 540bp 420bp,clip,width=0.48\textwidth]{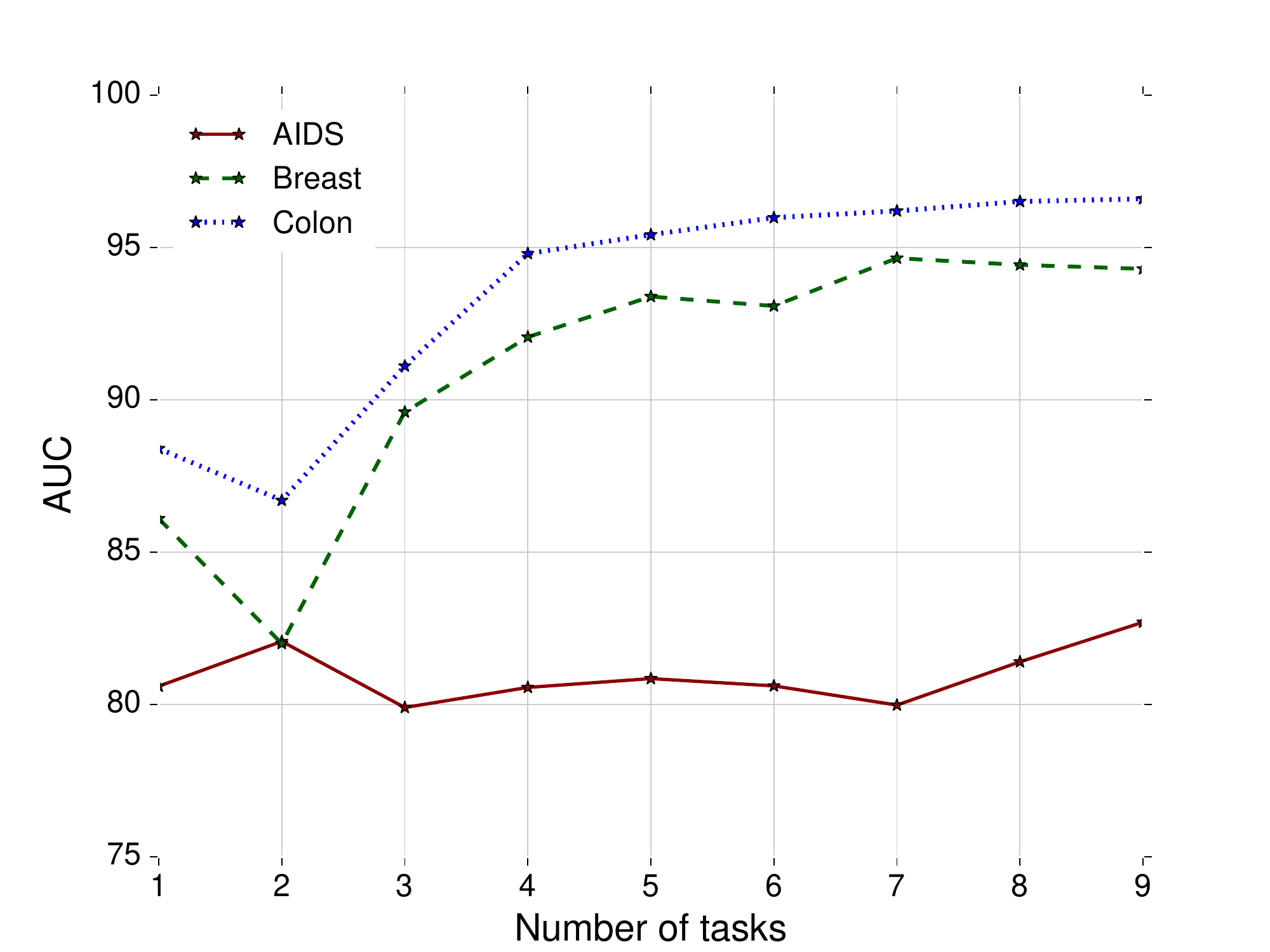} & \includegraphics[viewport=10bp 0bp 540bp 420bp,clip,width=0.48\textwidth]{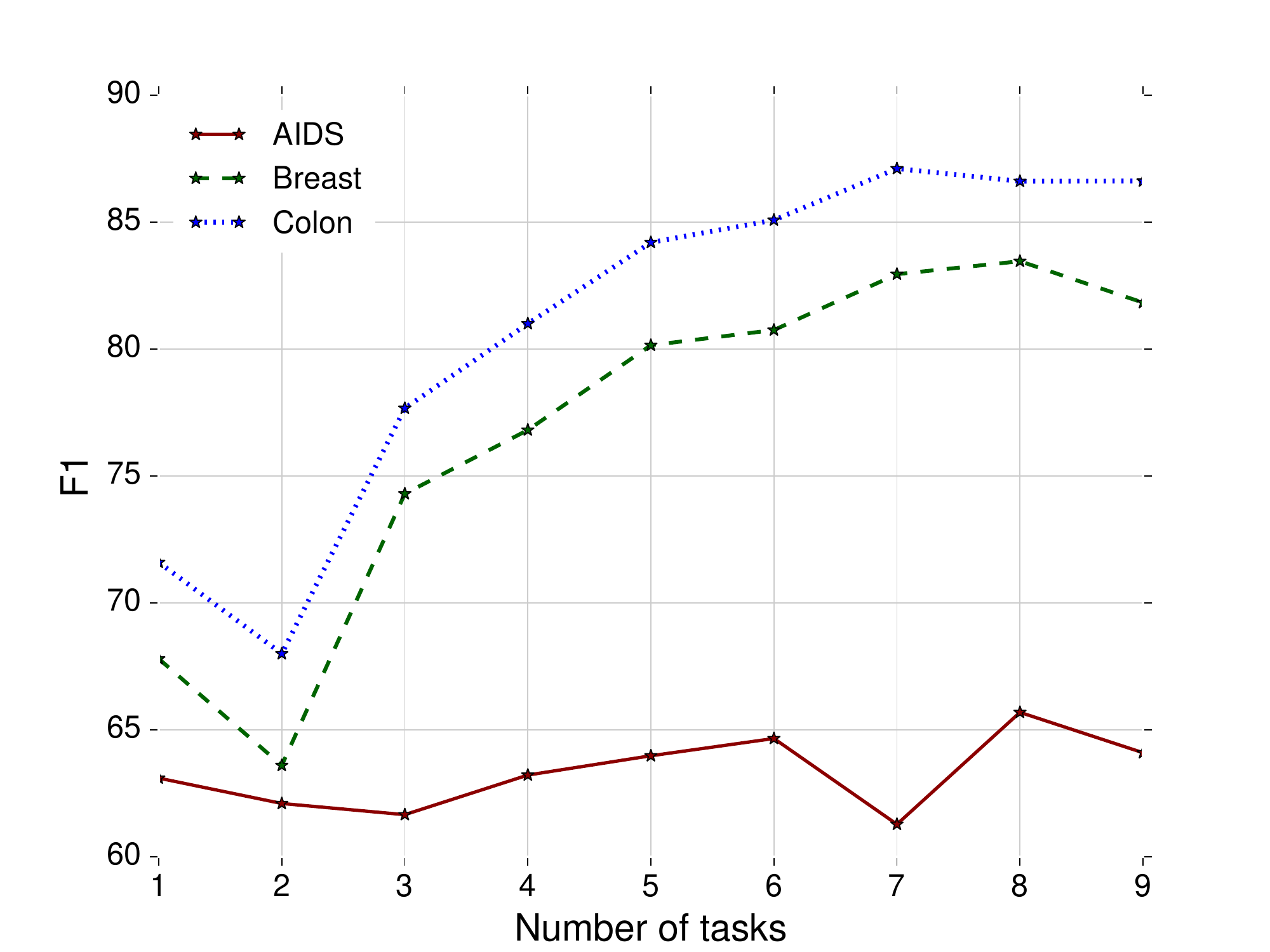}\tabularnewline
(a) & (b)\tabularnewline
\end{tabular}\caption{The performance of three datasets when increasing the number of jointly
training tasks, reported in (a) AUC and (b) F1-score.\label{fig:increasing_n_tasks}}
\end{figure}

Fig.~\ref{fig:increasing_n_tasks} illustrates the performance of
the three chosen datasets with different number of jointly training
tasks. The performance of Breast and Colon Cancer datasets decreases
when jointly trained with AIDS antiviral task, increases after adding
more tasks, and then remains steady or slightly reduces after seven
tasks. Joint training does not improve the performance on the AIDS
antiviral dataset.

\paragraph{Comparative results}

Table~\ref{tab:Results-all} reports results, measured in Micro F1-score,
Macro F1-score and the average AUC over all datasets. The best method
for separated training on fingerprint features is SVM with 66.4\%
of Micro F1-score and on graph structure is $\Model$ with the improvement
of 2.7\% over the non-structured classifiers. The joint learning settings
improve by 9.1\% of Micro F1-score and 10.7 \% of Macro F1-score gain
on fingerprint features and 8.7\% of Micro F1-score and 11.6\% of
Macro F1-score gain on graph structure.

\begin{table}
\centering{}\caption{Performance over all datasets, measured in Micro F1, Macro F1 and
the average AUC. \label{tab:Results-all}}
\begin{tabular}{cccc}
\hline 
Model & MicroF1 & MacroF1 & Average AUC\tabularnewline
\hline 
SVM & 66.4 & 67.9 & 85.1\tabularnewline
RF & 65.6 & 66.4 & 84.7\tabularnewline
GB & 65.8 & 66.9 & 83.7\tabularnewline
\hline 
NeuralFP \cite{duvenaud2015convolutional} & 68.2 & 67.6 & 85.9\tabularnewline
MT-NN \cite{ramsundar2015massively} & 75.5 & 78.6 & 90.4\tabularnewline
$\Model$ & \textbf{77.8} & \textbf{80.3} & \textbf{92.1}\tabularnewline
\hline 
\end{tabular}
\end{table}

\subsection{Chemical-chemical interaction (CCI) \label{subsec:Query:-Chemical-reaction}}

For this application we asks if two or more molecules interacts. Here
the queries can be about the strength of interaction given environmental
conditions (e.g., solution, temperature, pressure, presence of catalysts). 

\paragraph{Datasets}

We conducted experiments on chemical-chemical interaction data downloaded
from the STITCH database \cite{kuhn2007stitch} (Search Tool for InTeractions
of CHemicals), a network of nearly 1M chemical interactions for over
68K different chemicals. Each interaction between two chemicals has
confidence score from 0 to 999. Following \cite{kwon2018end}, we
extracted \emph{positive-900} (11,764 examples) and \emph{positive-800}
(92,998) from interactions with confidence scores more than 900 and
800, respectively and extract \emph{negative-0} (425,482 samples)
from interactions with confidence scores equal to zero. We then created
the CCI900 dataset from all the positive samples in positive-900 and
the same number of negative samples randomly drawn from negative-0.
CCI800 was also created similarly. Therefore, two datasets used for
the experiments - CCI900 and CCI800 have 23,528 and 185,990 samples,
respectively. The molecules were downloaded from the PubChem database
using CID (Compound ID) shown in STITCH. The tool RDKit\footnote{http://www.rdkit.org/}
was used to extract graph structures, atom and bond features, fingerprints
and SMILES of the molecules.

\subsubsection{Experiment settings}

We designed experiments on three different types of representations
of molecules: (i) fingerprint features, (ii) SMILES (Simplified Molecular-Input
Line-Entry System) - a string format to describe the structures of
molecules, and (iii) the graph structure information of the molecules. 

\paragraph{Baselines}

Fingerprint feature vectors are processed by Random Forests and Highway
Networks \cite{srivastava2015training}, which are two strong classifiers
in many applications \cite{pham2016faster}. Each data point consists
of two fingerprint vectors representing two molecules. We average
the two vectors as the input vector for the two baselines. SMILES
strings are modeled by DeepCCI \cite{kwon2018end}, a recent deep
neural network model for CCI. Each SMILES is represented by a matrix
where each row is a one-hot vector of a character. The matrix is then
passed through a convolution layer to learn the hidden representation
for each SMILE string. The two hidden vectors are then summed and
passed through a deep feedforward net to learn the final representation
of the interaction between the two molecules. We tune the hyper-parameters
for DeepCCI as suggested by the authors \cite{kwon2018end}.

\paragraph{Model setting}

We also conducted two other experiments to examine the effectiveness
of \emph{using side information as the query}. The attention read,
which is a weighted sum of all memory cells at the controller, might
not properly capture the global information of the graph while fingerprint
feature vectors or SMILES strings attain this information. Here, we
set two types of vectors as the queries: (i) the mean of two fingerprint
vectors and (ii) the hidden representation generated by DeepCCI. For
the latter setting, DeepCCI parameters are randomly initialized and
jointly learned with our model's parameters. 

\subsubsection{Results}

\paragraph{The effect of multiple attentions}

\begin{figure*}
\begin{centering}
\begin{tabular}{cc}
\includegraphics[viewport=0bp 0bp 530bp 430bp,clip,width=0.48\textwidth]{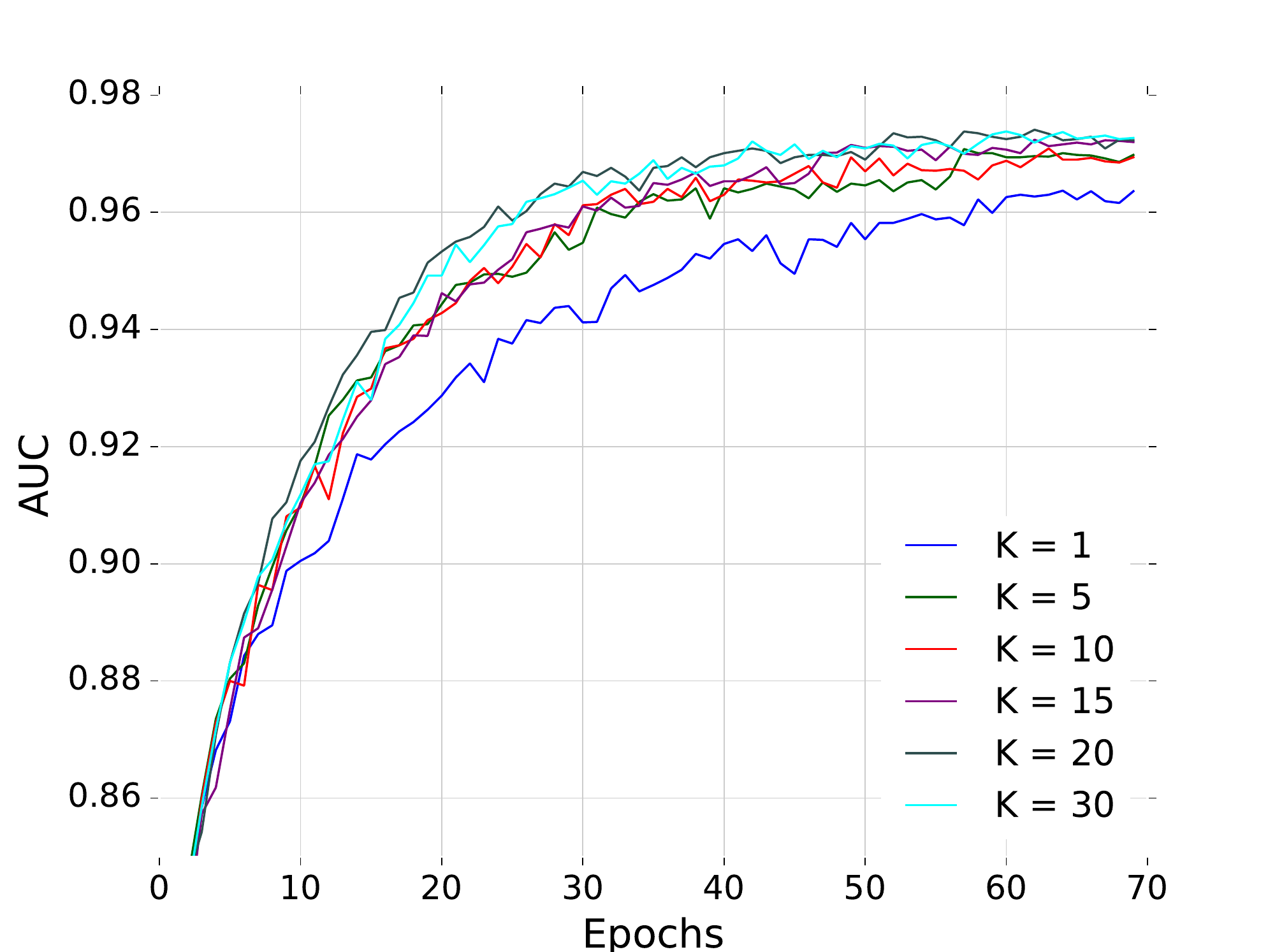} & \includegraphics[viewport=0bp 0bp 530bp 430bp,clip,width=0.48\textwidth]{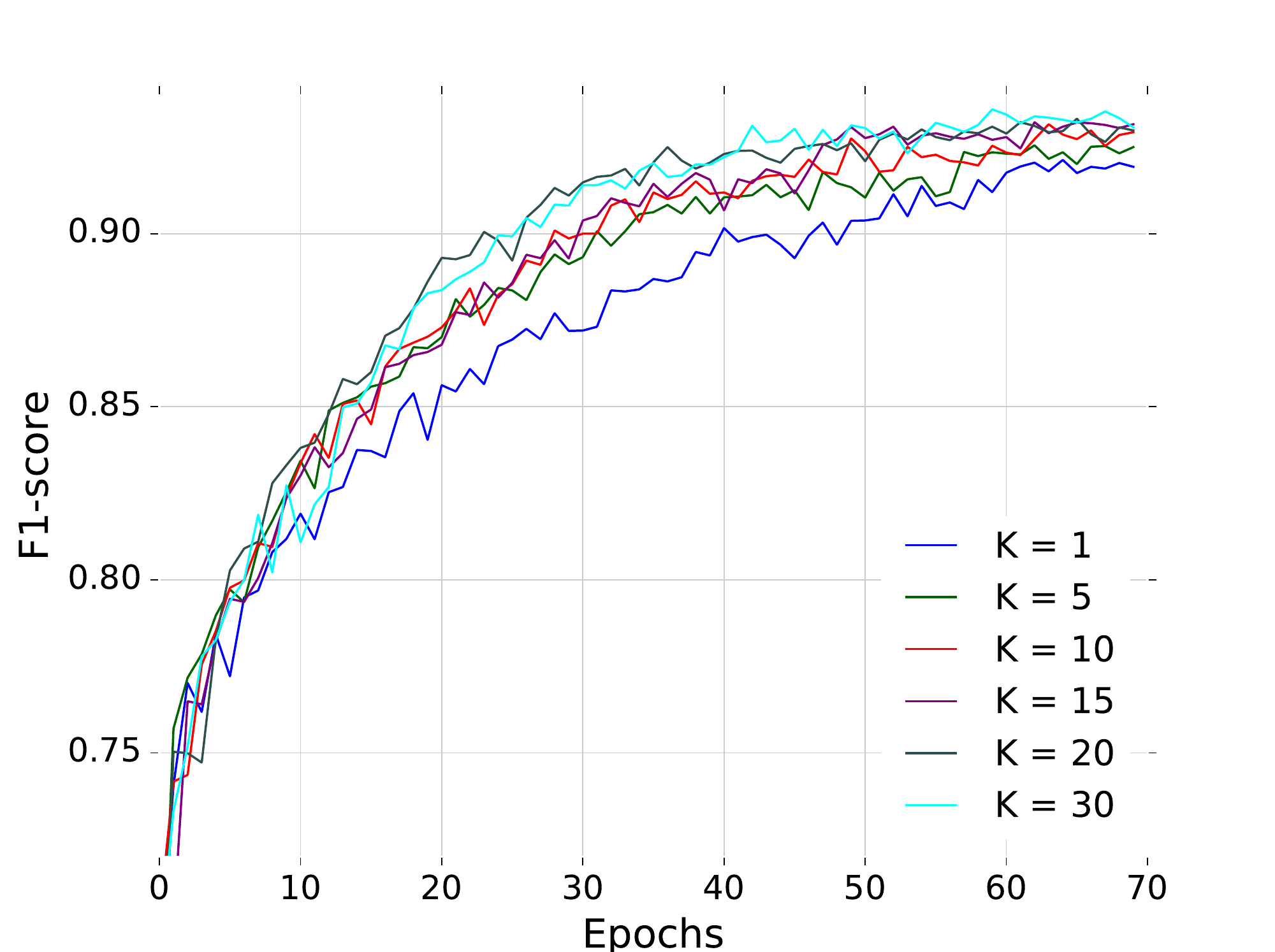}\tabularnewline
(a)  & \tabularnewline
\end{tabular}
\par\end{centering}
\caption{The performance of $\protect\Model$ with different number of reading
heads $K$ during training, reported in (a) AUC and (b) F1-score.
Best viewed in color. \label{fig:DeepGI-multiAtt-results}}
\end{figure*}

In CCI, substructures from different molecules interact, leading to
multiple substructure interactions. In our model, we expect that each
attention reading head is able to capture a specific interaction of
substructures. Hence, an appropriate number of attention heads can
capture all substructure interactions and provide more abundant information
about the graph interaction, which may improve the prediction performance.
Here, we evaluated the improvement in prediction brought by the number
of attention heads $K$. Taking $\Model$ without the side information
query, we varied $K$ by 1, 5, 10, 15, 20 and 30, and trained the
resulting six models independently on the CCI900 dataset. Fig.~\ref{fig:DeepGI-multiAtt-results}
reports the performance changes during training for different $K$.
We can see that when $K>1$, increasing $K$ only slightly improves
the performance while there is a bigger gap between the performance
of $K=1$ and $K>1$. There is not much difference when $K$ is increased
from 20 to 30. It is possible that when the number of attention heads
is large, they collect similar information from the graphs, leading
to saturation in performance.

\paragraph{The effect of the side information as query}

\begin{figure*}
\centering{}\includegraphics[viewport=10bp 20bp 650bp 415bp,clip,width=0.8\textwidth]{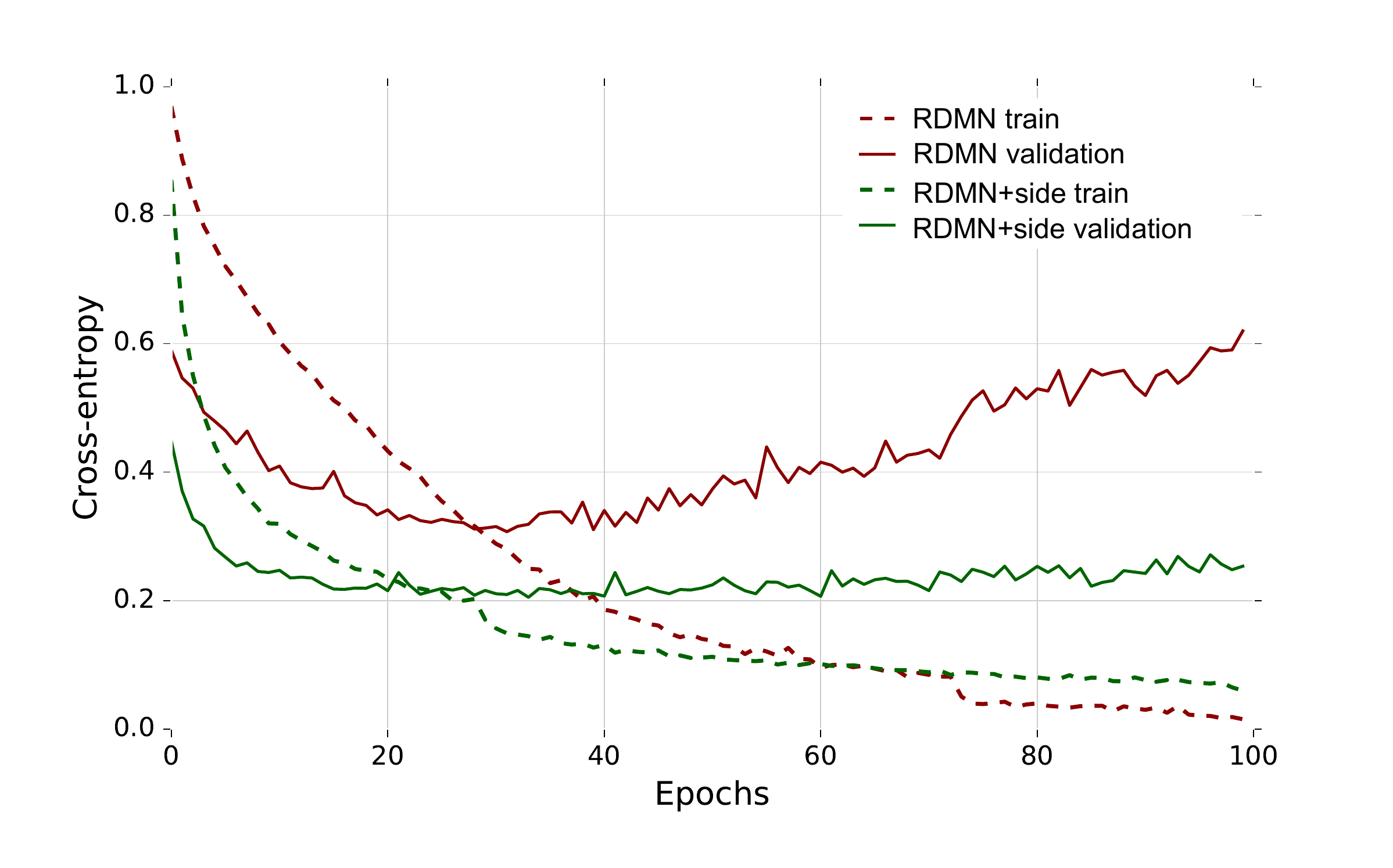}\caption{Training and validation losses of $\protect\Model$ with and without
side information during training. Best viewed in color. \label{fig:DeepGI-side-info}}
\end{figure*}

While the neighborhood aggregation operation in our model can effectively
capture the substructure information, the weighted sum of all memory
cells in the attention mechanism might not properly collect the global
information of the whole graph. Hence, using the side information
containing the global information such as Fingerprints or SMILES strings
might help the training. We have shown in Table~\ref{tab:Chemical-reaction-results}
that using fingerprint vectors or the hidden state generated by DeepCCI
as the query for our model can improve the performance on the both
datasets. We also found that even though using side information query
increases the number of training parameters, it can prevent overfitting.
Fig.~\ref{fig:DeepGI-side-info} shows the loss curves on the training
and the validation set of $\Model$ with a single attention head when
the query is set as constant and when the query is the hidden stated
produced by DeepCCI model ($\Model$+SMILES). We can see from the
figure that the validation loss of $\Model$ starts to rise quickly
after 40 epochs while the validation loss of $\Model$+SMILES still
remains after 100 epochs.

\paragraph{Comparative results}

Table~\ref{tab:Chemical-reaction-results} reports the performance
of the baselines and our proposed model in different settings on the
two dataset CCI900 and CCI800 reported in AUC and F1-score. SMILES
features with DeepCCI outperformed the fingerprint features with Highway
Networks by 3.8\% F1-score on CCI900 and by 3.6\% F1-score on CCI800.
Our model with a single attention head is slightly better than DeepCCI
on both datasets. Interestingly, using multiple attention heads and
side information (fingerprint and SMILES features) as the query both
help to improve the performance. For CCI900, $\Model$ with SMILES
query improves 1.7\% F1-score on single attention setting and 1.2\%
on multiple attention settings.

\begin{table}[h]
\centering{}%
\begin{tabular}{ccccc}
\hline 
\multirow{2}{*}{} & \multicolumn{2}{c}{CCI900} & \multicolumn{2}{c}{CCI800}\tabularnewline
\cline{2-5} 
 & AUC & F1-score & AUC & F1-score\tabularnewline
\hline 
Random Forests & 94.3 & 86.4 & 98.2 & 94.1\tabularnewline
Highway Networks & 94.7 & 88.4 & 98.5 & 94.7\tabularnewline
DeepCCI \cite{kwon2018end} & 96.5 & 92.2 & 99.1 & 97.3\tabularnewline
\hline 
$\Model$ & 96.6 & 92.6 & 99.1 & 97.4\tabularnewline
$\Model$+multiAtt & 97.3 & 93.4 & 99.1 & 97.8\tabularnewline
\hline 
$\Model$+FP & 97.8  & 93.3 & 99.4 & 98.0\tabularnewline
$\Model$+multiAtt+FP & 98.0 & 94.1 & 99.5 & 98.1\tabularnewline
\hline 
$\Model$+SMILES & 98.1 & 94.3 & 99.7 & 97.8\tabularnewline
$\Model$+multiAtt+SMILES & \textbf{98.1} & \textbf{94.6} & \textbf{99.8} & \textbf{98.3}\tabularnewline
\hline 
\end{tabular}\caption{The performance on the CCI datasets reported in AUC and F1-score.
\emph{FP} stands for fingerprint and \emph{multiAtt} stands for multiple
attentions. \label{tab:Chemical-reaction-results}}
\end{table}

\section{Related work \label{sec:Related-work}}

\paragraph{Working memory and reasoning}

This paper is partly inspired by the concept of ``working memory'',
a brain faculty for holding and manipulating information for an extended
period of time (e.g., seconds to minutes) \cite{baddeley1992working}.
Working memory is a critical component for high-level cognition (e.g.,
reasoning, meta-reasoning) \cite{diamond2013executive}. Observations
found that working memory arises through functional coordination between
brain regions \cite{stokes2015activity}, which may suggest the graph-theoretical
approach for cognitive modeling. An important feature of working
memory is the fast loading (or binding) of information into the memory,
allowing rapid task-switching and selective attention \cite{bao2007binding}.
Our work draws inspiration from these findings and theory, but does
not aim to be biologically relevant. Rather we aim for building a
neural network capable of \emph{neural reasoning} \cite{jaeger2016artificial}.
This capability is needed for tasks such as graph traversal \cite{graves2016hybrid},
knowledge graph completion \cite{socher2013reasoning} and question
answering \cite{sukhbaatar2015end,xiong2017dynamic}. To facilitate
relational reasoning, recent neural networks \cite{santoro2017simple}
have been proposed to model pair-wise relationship between objects.
A recurrent variant allows more dynamic reasoning process \cite{santoro2018relational}.
Our work contributes to this line of research by introducing a controller
into a relational memory module, and thus helping in answering arbitrary
queries. 

\paragraph{Graph representation and querying}

There has been a surge of interest in learning graph representation
for the past few years \cite{bruna2014spectral,li2016gated,niepert2016learning,ying2018hierarchical}.
Deep spectral methods have been introduced for graphs of a given
adjacency matrix \cite{bruna2014spectral}, whereas we allow arbitrary
graph structures, one per graph. Several other methods extend convolutional
operations to irregular local neighborhoods \cite{atwood2016diffusion,niepert2016learning,pham2017column},
or employ recurrent paths along the random walk from a node \cite{scarselli2009graph}.
Graph dynamics has been recently studied, e.g., Gated Graph Transformer
\cite{johnson2017learning}, which models graph state transition over
time. Related to but distinct from our work is the work about querying
on graphs in the database community \cite{libkin2016querying}. There
are sub-problems that can benefit from $\OurModel$, however, including
graph matching, maximum clique and segmentation (clustering). Our
work adds to the existing literature by allowing arbitrary querying
from a single or multiple graphs. Note that our system can query about
inferred knowledge instead of just factual knowledge.

\paragraph{Memory-augmented neural nets (MANN)}

MANNs are new developments but have found a wide range of applications,
including question-answering \cite{kumar2016ask,sukhbaatar2015end},
graph processing \cite{graves2016hybrid}, algorithmic tasks \cite{graves2016hybrid},
meta-learning \cite{santoro2016meta}, healthcare \cite{le2018dual}
and dialog systems \cite{le2018variational}. Notable MANN architectures
include Neural Turing Machine \cite{graves2014neural} and its recent
cousin, the Differentiable Neural Computer \cite{graves2016hybrid}.
Here all operations are fully differentiable allowing end-to-end training
with and the memory dynamically updated during the reasoning process.
This design is powerful but poses great challenges for implementation.
End-to-End Memory Networks \cite{sukhbaatar2015end} simplify this
by loading the entire input matrix into the memory and fixing the
memory content. Similar architectures have been subsequently proposed
to solve specific tasks, including the Dynamic Memory Networks for
question-answering problems \cite{kumar2016ask}, the Recurrent Entity
Net \cite{henaff2017tracking} for tracking entities, and the key-value
structure for rare-events \cite{kaiser2017learning}. Limited work
has been done for the structured memory \cite{bansal2017relnet,parisotto2017neural}.
Our $\Model$ differs from these models by using a dynamic memory
organized as a graph of cells. The cells interact not only with the
controller but also with other cells to embed the substructure in
their states. A potential mechanism for a dynamic working memory is
introduced in \cite{do2018learning,santoro2018relational}, where
relations between cells are either learned \cite{do2018learning}
or dynamically estimated through self-attention \cite{santoro2018relational}.
Our work builds on top of this memory module by introducing a controller
to manage update of the memory matrix. 

\paragraph{Source code modeling}

Our application of $\Model$ for software is in line with the current
direction of modeling source code using structures such as trees \cite{dam2018deep}
and graphs \cite{allamanis2018learning}. The recurrent nature of
the $\Model$, in principle, would capture a program execution without
actually running the code \cite{choi2017end}.

\paragraph{Molecular graphs modeling}

Our application to chemical compound classification bears some similarity
to the work of \cite{duvenaud2015convolutional}, where graph embedding
is also collected from node embedding at each layer and refined iteratively
from the bottom to the top layers. However, our treatment is more
principled and more widely applicable to multi-typed edges \cite{coley2017convolutional,kearnes2016molecular}.
The controller used in $\Model$ resembles the virtual atom mentioned
in \cite{gilmer2017neural} for a single molecule setting. Our application
to chemical-chemical interaction (CCI) differs radically from existing
machine learning work in the area, whose prediction pipeline typically
consists of manual feature engineering before a classifier is applied
(e.g., see \cite{kwon2018end} for references therein). To the best
of our knowledge, DeepCCI \cite{kwon2018end} is the most recent work
leveraging the end-to-end learning capability of neural networks for
CCI. DeepCCI employs string representation of molecules, then uses
CNN coupled with a pairwise loss defined by a Siamese network. In
contrast, our approach using $\Model$ models the molecular graphs
and their complex interaction directly. $\Model$ is theoretically
powerful as it can model interaction of multiple compounds. Related
but distinct from CCI is the problem of predicting chemical reaction,
where the goal is to produce reaction products, which are essentially
predicting bond changes (e.g., see \cite{do2019graph} for references
therein).

\section{Conclusion \label{sec:Discussion}}

We have proposed $\OurModel$ (which stands for Relational Dynamic
Memory Network), a new neural network augmented with a graph-structured
``working'' memory. $\Model$ supports arbitrary querying over a
set of graph-structured objects, that is, solving the problem of the
form \texttt{(query, \{set of graphs\}, ?)}. We applied $\Model$
to three tasks: software source code vulnerability, molecular bioactivity
and chemical-chemical interaction. The results are competitive against
rivals, demonstrating a possibility to implement differentiable neural
reasoning for highly complex tasks.

A limitation of the current formulation is that the memory structure
in $\Model$ is fixed once constructed from data graphs. A future
work would be deriving dynamic memory graphs that evolve with time,
e.g., using the technique introduced in \cite{ying2018hierarchical},
or a gated message passing between memory cells. Also, multiple controllers
may be utilized to handle asynchronous issues (e.g., see \cite{le2018dual})
as they can partially share some memory components. Another limitation
of the current work is that the model was only validated on single
graphs and pairwise graph-graph interaction, but $\Model$ does not
have such intrinsic restriction. For example, it would be interesting
to see how $\Model$ performs on high-order compound interactions
-- a setting of paramount importance in estimating treatments effects
\cite{weinstein2017prediction}.

\bibliographystyle{plain}

\end{document}